\documentclass[sigconf,9pt]{acmart}

\usepackage{booktabs} 
\usepackage{color}

\setcopyright{none}

\acmConference[FAT/ML '17]{2017 Workshop on Fairness, Accountability, and Transparency in Machine Learning}{August 2017}{Halifax, Canada}
\acmYear{2017}
\copyrightyear{2017}
\acmDOI{}
\acmISBN{}

\acmPrice{}

\begin{document}

\title{Data Decisions and Theoretical Implications when\\Adversarially Learning Fair Representations}
\acmTitle{Data Decisions and Theoretical Implications when Adversarially Learning Fair Representations}

\author{Alex Beutel, Jilin Chen, Zhe Zhao, Ed H. Chi}
\affiliation{
  \institution{Google Research}
  \city{Mountain View}
  \state{CA}
\\\{alexbeutel, jilinc, zhezhao, edchi\}@google.com
}

\begin{abstract}
How can we learn a classifier that is ``fair'' for a protected or sensitive group, when we do not know if the input to the classifier belongs to the protected group?  How can we train such a classifier when data on the protected group is difficult to attain?  In many settings, finding out the sensitive input attribute can be prohibitively expensive even during model training, and sometimes impossible during model serving.  For example, in recommender systems, if we want to predict if a user will click on a given recommendation, we often do not know many attributes of the user, e.g., race or age, and many attributes of the content are hard to determine, e.g., the language or topic.  Thus, it is not feasible to use a different classifier calibrated based on knowledge of the sensitive attribute.

Here, we use an adversarial training procedure to remove information about the sensitive attribute from the latent representation learned by a neural network.  
In particular, we study how the choice of data for the adversarial training effects the resulting fairness properties.  We find two interesting results: a small amount of data is needed to train these adversarial models, and the data distribution empirically drives the adversary's notion of fairness.
\end{abstract}

\maketitle

\section{Introduction}

In recent years, researchers have recognized unfairness in ML models as a significant issue.  In numerous cases, machine learned models offer much worse quality results for a protected group than for the population overall.
The problem has garnered significant attention in the research community, with some working to define and understand ``fairness,'' and others working to develop techniques to ``de-bias'' ML algorithms and models.

One commonly understood source of bias is skewed data---for example, when a group of users is underrepresented in the training data and, as a result, the model is less accurate for that group \cite{beutel2017beyond,bolukbasi2016man}.  However, much of the recent work on de-biasing models ignores the implications of the data used to perform the de-biasing.

We consider the case where it is difficult or expensive to find out if a datapoint is from the protected group or not, i.e., to get labels of the \emph{sensitive attribute}.  This is common in many cases, such as when the sensitive attribute is private, such as personal information about a user, or when the sensitive attribute is in some way imprecisely defined, such as what topic a piece of user generated content is about.  The scarcity of data can be further exacerbated by the underlying skew in data distribution.  For example, if only 5\% of examples belong to the protected class, it would require labeling a much larger random sample of data in order to have a large dataset for both the protected class and the general population.  

There are two significant implications of this constraint.
First, during model training, any method used to de-bias the underlying model or learn a ``fair'' model must account for the limited and often skewed data about the bias, lest the de-biasing algorithm fall victim to the same issues as the original model.  
While a few model structures have been proposed that are related to the approach we take here, they do not study or discuss the effect of limited training data \cite{bousmalis2016domain,ganin2016domain}.

Second, after the de-biased model is trained and when it is applied as a predictor for unlabeled data, it cannot rely on knowing if the example in question is from the protected class or not.  Recent literature sharpening the definition of fairness has relied on a calibration procedure that breaks this constraint \cite{hardt2016equality,kleinberg2016inherent}.

In this work, we explore both of these problems by using adversarial learning to de-bias latent representations. That is, we build a multi-head deep neural network (DNN) where the model is trying to predict the target class with one head while simultaneously preventing the second head from being able to accurately predict the sensitive attribute.
With this approach,  we make the following contributions:
\begin{enumerate}
    \item We connect theoretically the different definitions of fairness with the adversarial training objective and the choice of dataset used for the adversary.
    \item We explore empirically how much data is needed to effectively de-bias a learned ML model.
    \item We study empirically how different data distributions use in the adversarial learning effect the resulting fairness of the model.
\end{enumerate}

\section{Related Work}
\paragraph{Fairness Definitions}
As fairness in machine learning has become a societal focus, researchers have tried to develop useful definitions of ``fairness'' in machine learning systems.  Notably, Hardt et al.\ and Kleinberg et al. ~\cite{hardt2016equality,kleinberg2016inherent} have both offered novel theoretical work explaining the trade-offs between demographic parity, previously focused on as ``fair,'' and alternative formulations focused more closely on model accuracy.  We will primarily work off of the definitions offered in~\cite{hardt2016equality}.

Along with the theoretical underpinnings, Hardt et al.~\cite{hardt2016equality} offers a method for achieving equality of opportunity, but does so through a post-processing algorithm, taking as input the model's prediction and the sensitive attribute.  Kleinberg et al.~\cite{kleinberg2016inherent} likewise offers a calibration technique to achieve fairness.  These approaches are also problematic in many cases when the sensitive attribute is not observable at inference time.

\paragraph{Fair-er Machine Learning}
A growing body of literature is aimed at improving model performance for underserved parts of the data.  For example, Beutel et al.~\cite{beutel2017beyond} uses hyperparameter optimization to improve model performance for underserved regions of the data in collaborative filtering.  More directly in the ``fairness'' literature, Zemel et al.~\cite{zemel2013learning} first attempted to learn ``fair'' latent representations by directly enforcing statistical parity during unsupervised learning.

\paragraph{Adversarial Training}
Combining competing tasks has been found to be a useful tool in deep learning.  In particular, researchers have included an adversary to help compensate for skewed data distributions in domain adaptation problems for robotics and simulations~\cite{bousmalis2016domain,ganin2016domain}.  Researchers have also applied similar techniques for making models fair by trying to prevent biased latent representations~\cite{edwards2015censoring,louizos2015variational}.  This literature has generally not been as precise in terms of which definition of fairness they are optimizing for and what data is used for the adversarial objective.  If the definition is mentioned at all, the work often focuses on demographic parity, which, as Hardt et al.~\cite{hardt2016equality} explains, has many drawbacks.  We explore the intersection of these research efforts.

\begin{figure}[tb]
    \centering
    \includegraphics[width=0.95\columnwidth]{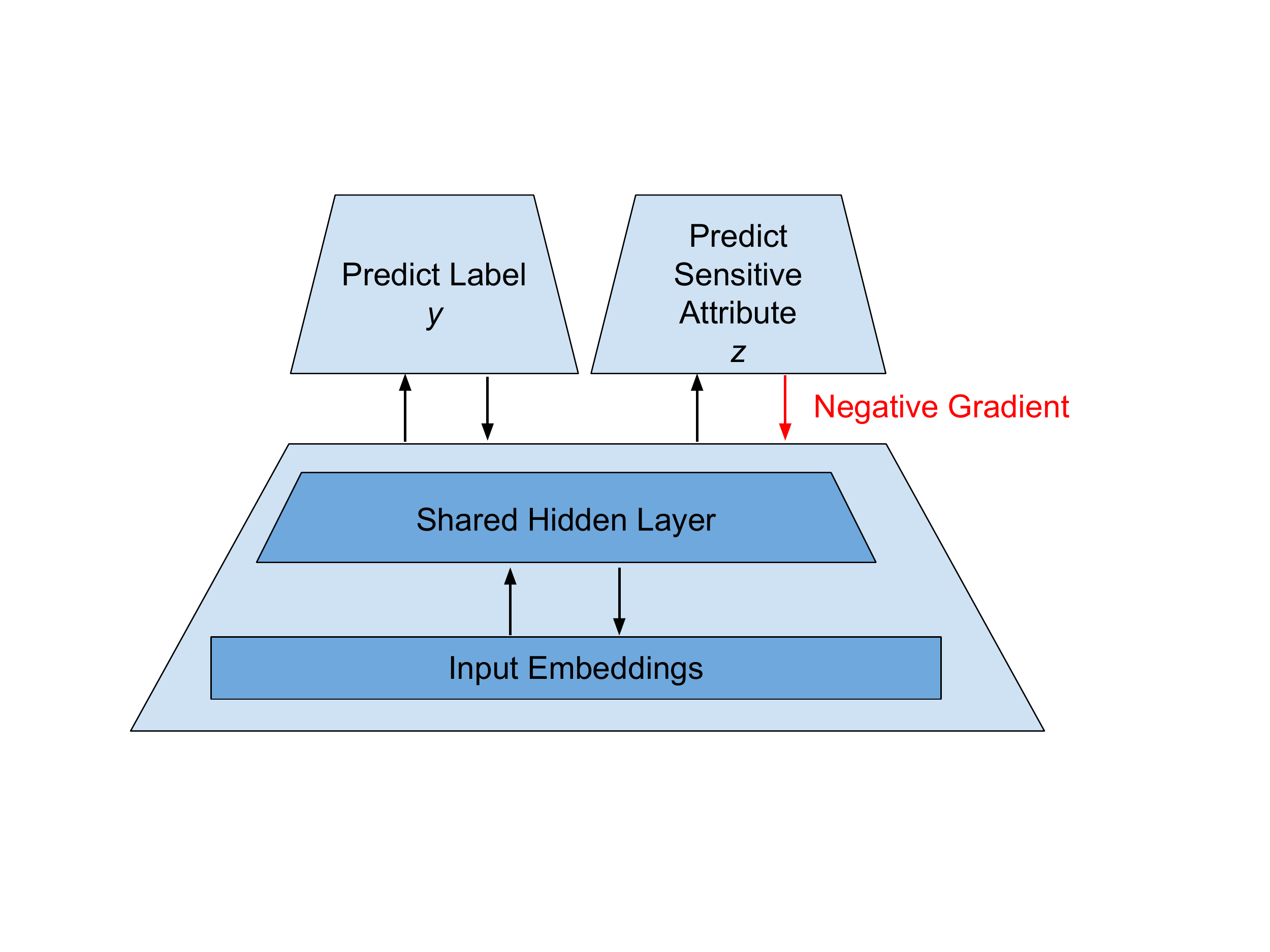}
    \vspace{-12mm}
    \caption{Model for adversarial training.}
    \label{fig:model}
\end{figure}

\section{Model Structure and Learning}
The adversarial training procedure described here is closely related to Edwards and Storkey~\cite{edwards2015censoring}, but we describe it here for completeness and to explain in detail how our learning procedure differs from classic approaches.

\subsection{Model Structure}
Our primary task is given input $X$ to predict some label $Y$.  In this case $Y$ can be either real or categorical.  We assume that we have a model of the form $Y=f(g(X))$ where $g(X)$ produces an embedding $h$ and $f(h)$ produces a prediction $Y$.  Note here $f$ and $g$ can be arbitrary neural networks with parameters learned through typical back propagation.
 
We assume that, for each example, there exists a feature $Z$ that we consider to be sensitive or protected, and for which we want our predictions to be independent of this feature.  Importantly, even if the feature $Z$ is not used as an input to $g$, it may be correlated with other observed features.
 
Additionally, we assume that we can observe $Z$ for some subset of $X$, and let's call this set $S$.  We then train a second adversarial classifier $a(g(S)) = Z$.  Note that $g$ is the same function as above, but $a(h)$ is a new function that predicts $Z$, given the hidden embedding $h$.

\subsection{Learning Algorithm}
Our goal is for $f(g(X))$ to predict $Y$ and $a(h)$ to predict $Z$ as well as possible, but for $g()$ to make it hard for adversary $a()$ to predict $Z$.  To be more precise, we assume we have a normal loss $L_Y(f(g(X)), Y)$ for predicting $Y$, such as cross entropy loss for classification or squared error for regression.  We also assume we have a cross entropy classification loss $L_Z(a(g(S)), Z)$ for predicting $Z$.  
 
However, if we were to minimize $L_Y + L_Z$, then $g(X)$ would be encouraged to predict $Z$, rather than discouraged.  As such, we make the following change: $L_Z(a(J_\lambda(g(S))), Z)$.  Here $J_\lambda$ is an identity function with a negative gradient.  That is, $J(g(S)) = g(S)$ and  $\frac{dJ}{dS} = -\lambda \frac{d g(S)}{dS}$.   As a result, while $a()$ is trained to minimize the classification error, $g()$ is trained to maximize the classification error for $Z$.
Therefore, $g()$ is trained from $L_Y$ to predict $Y$ and from $L_Z$ to not encode any information allowing the model to predict $Z$.
$\lambda$ determines the trade-off between accuracy and removing information about sensitive attribute $Z$.
 
As such, we train our model with the following objective:
\begin{align}
    \min \left[\sum_{(x,y) \in X} L_Y(f(g(x)), y) + \sum_{(x,z) \in S} L_Z(a(J_\lambda(g(x))), z)\right]
\end{align}

\section{Data Selection \& Fairness Definition}
\label{sec:datadef}

One key point that is often overlooked is the properties of dataset $S$.
Because obtaining $S$ can be difficult, we ask: what are the implications of the distribution of $S$ over $Y$ and $Z$?  Interestingly, we find that the distribution over $Y$ corresponds to different definitions of fairness.
In explaining this connection, we consider a hypothetical example of a model trained to predict if a piece of content is ``dangerous'' $Y$, but would like to avoid biasing by topic $Z$.  
 
If the adversarial head of our model uses data $S$ that contains both $Y=1$ and $Y=0$, then the model will be encouraged to \emph{never} encode information about the sensitive attribute $Z$, e.g. the topic.  That is, latent representation $h$ would be uncorrelated with $Z$.  One result of that is that the probability of predicting whether the content is dangerous $\hat{Y}$ is independent of topic $Z$; that is $P(\hat{Y} = 1 | Z = 1) = P(\hat{Y} = 1 | Z=0)$.  This independence between prediction $\hat{Y}$ and sensitive attribute $Z$ is known as \emph{demographic parity}.

In contrast, consider the case of the adversarial head of our model only using data for $Y=1$ (not dangerous content).  In that case, the model is trained to not encode information about the topic $Z$ \emph{only when the content is not dangerous} $(Y=1)$.  Note, this means the model can still encode topic-specific features for why content could be dangerous, such as specific hate slurs.

Probabilistically, this can be stated as $h$ should be uncorrelated with topic $Z$ when the content is not dangerous $Y=1$. As a result, the probability of predicting whether the content is dangerous $\hat{Y}$ is independent of $Z$ conditioned on the content actually being not dangerous $Y=1$.  Mathematically, that is $P(\hat{Y} = 1| Y=1, Z=1) = P(\hat{Y} = 1| Y=1, Z=0)$.  Interestingly, this is precisely Hardt et al.'s \emph{equality of opportunity} \cite{hardt2016equality}.

Finally, we can enforce the reciprocal equality of opportunity statement.  If the adversarial head is only trained on data for dangerous content $Y=0$, the model is encouraged to predict that dangerous content is no more or less likely to be dangerous based on its topic.
Mathematically, that is $P(\hat{Y} = 0| Y=0, Z=1) = P(\hat{Y} = 0| Y=0, Z=0)$.
This is still equality of opportunity but for the negative class $Y=0$. 

Given these theoretical connections, we now consider: how do these different training procedures effect our models in practice?

\section{Experiments}
We now explore empirically what are the effects of using different data distributions for the adversarial head of our model and observe whether the experimental results align with the theoretical connections described in Section \ref{sec:datadef}.

\paragraph{Data}
We run experiments on the Adult dataset from the UCI repository \cite{Lichman:2013}.  Here, we try to predict whether a person's income is above or below \$50,000 and we consider the person's gender to be a sensitive attribute.  The dataset is skewed with 67\% men.  One interesting property of this dataset is that there isn't demographic parity in the underlying data: 30.6\% of men in the dataset made over \$50,000, but only 16.5\% of women did as well.  Additionally, only 15\% of the people making over \$50,000 were female.  The complete breakdown is shown in Table \ref{tab:dataset}.  Results are reported on a test set of 8140 users.

\paragraph{Model} We train a model with a single 128-width ReLU hidden layer.  Both the adversarial head and the primary head are trained with a logistic loss function, and we use the Adagrad \cite{duchi2011adaptive} optimizer in Tensorflow with step size 0.01 for 100,000 steps.  We use a batch size of 32 for both heads of the model.  Each head uses a different input stream so that we can vary the data distribution for the two heads separately.  In each experiment, we run the training procedure 10 times and average the results.  Each model's classification threshold is calibrated to match the overall distribution in the training data.

\begin{table}[tb]
    \centering
    \begin{tabular}{ccc}
         & Male & Female  \\
        $\leq50$K & 15,128 & 9592 \\
        $>50$K & 6662 & 1179 \\
    \end{tabular}
    \vspace{3mm}
    \caption{Dataset breakdown by gender and income.}
    \vspace{-9mm}
    \label{tab:dataset}
\end{table}

\paragraph{Metrics}
In addition to the typical accuracy, we will track two measures used in the fairness literature.  To understand the demographic parity, we will track:
\begin{align}
    ProbTrue_z = P(\hat{Y}=1|Z=z) = \frac{TP_z + FP_z}{N_z}
\end{align}
Here $N_z$ is the number of examples with sensitive attribute set to $z$, and $TP_z$ and $FP_z$ are the number of true positive and false positives in class $z$, respectively.  
To understand the equality of opportunity we measure:
\begin{align}
    ProbCorrect_{1,z} = P(\hat{Y}=1|Z=z, Y=1) = \frac{TP_z}{TP_z + FN_z}\\
    ProbCorrect_{0,z} = P(\hat{Y}=0|Z=z, Y=0) = \frac{TN_z}{TN_z + FP_z}
\end{align}
With these two terms, we define our two metrics of fairness:
\begin{align}
    {\it Parity\;Gap} &= |ProbTrue_1 - ProbTrue_{0}|\\
    {\it Equality\;Gap}_y &= |ProbCorrect_{y,1} - ProbCorrect_{y,0}|
\end{align}
Note, for both of these measures, lower is better.

\paragraph{Experimental Variants}
We explore a few different variants of training procedures to understand the impact of training data for the adversarial head on accuracy and model fairness.  In particular, we vary the distribution over sensitive attribute $Z$, the distribution over target $Y$, and the size of the data.   We test with two different distributions over $Z$: (1) \textbf{unbalanced:} the distribution over $Z$ matches the distribution of the overall training data, (2) \textbf{balanced:} an equal number of examples with $Z=1$ and $Z=0$.  
We consider three distributions over $Y$:
(1) \textbf{low income:} only use examples for $Y=0$ ($\leq50$K), (2) \textbf{high income:} only use examples for $Y=1$ ($>50$K), and (3) use an equal number of examples with $Y=1$ and $Y=0$.  Last, we use datasets of size in $\{500, 1000, 2000, 4000\}$ examples; when unspecified, we are using the dataset with 2000 examples.  Because the adversarial dataset is much smaller than general training dataset, we will reuse data in the adversarial dataset at a much higher rate than the general dataset.  Finally, in each experiment, we vary the adversarial weight $\lambda$ and observe the effect on metrics.

\paragraph{Baseline}
We consider as a baseline the performance when there is no adversarial head.
There, we observe an accuracy of 0.8233, ${\it Equality}_{\leq50K} = 0.1076$, ${\it Equality}_{>50K} = 0.0589$, and ${\it Parity} = 0.1911$.  
The experiments below primarily decrease accuracy but improve fairness; as we are primarily interested in the relative effects of the different adversarial modeling choices, we do not repeat the baseline results below.

\begin{figure*}[tbh]
    \centering
    \includegraphics[width=0.49\columnwidth]{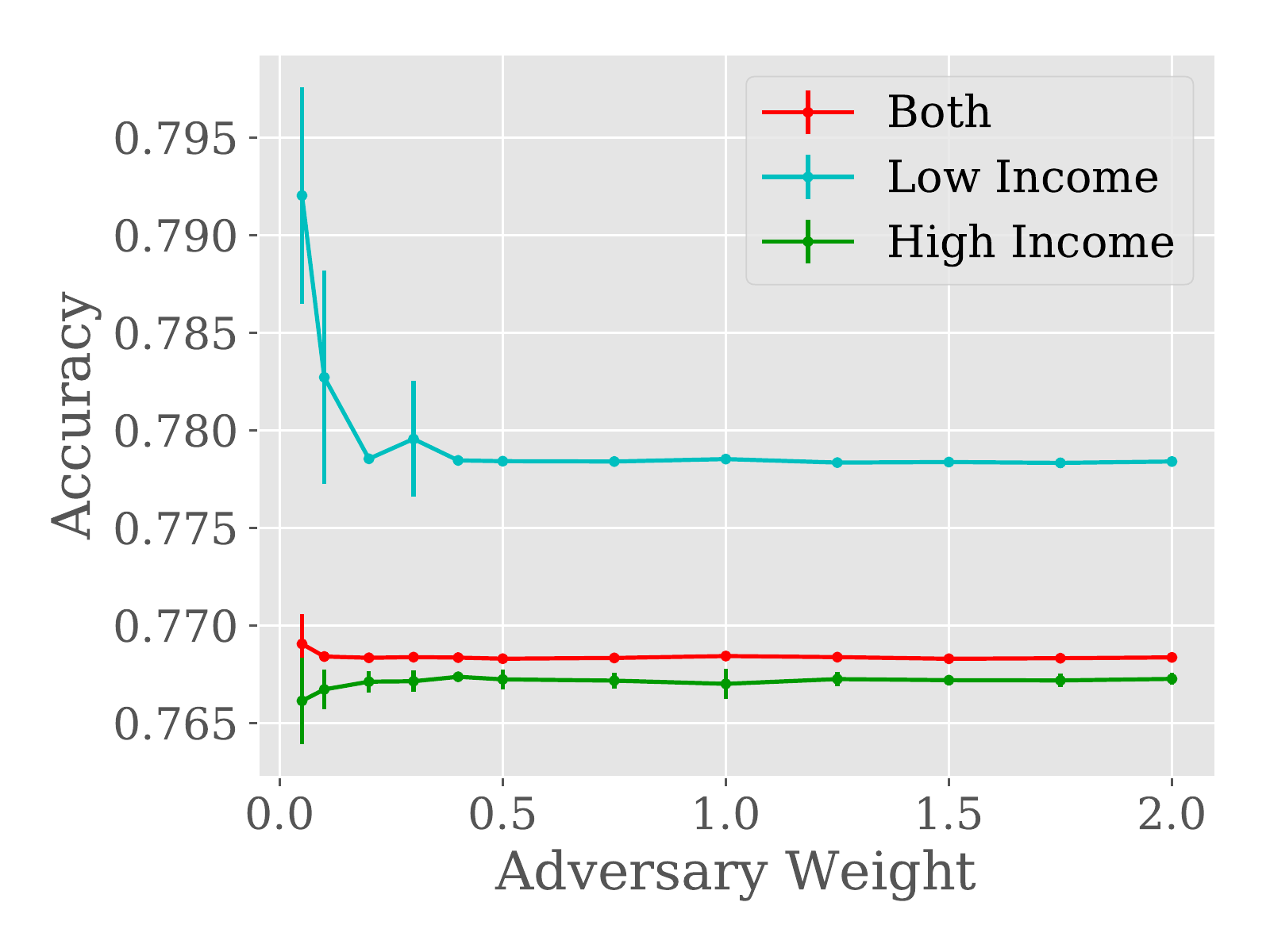}
    \includegraphics[width=0.49\columnwidth]{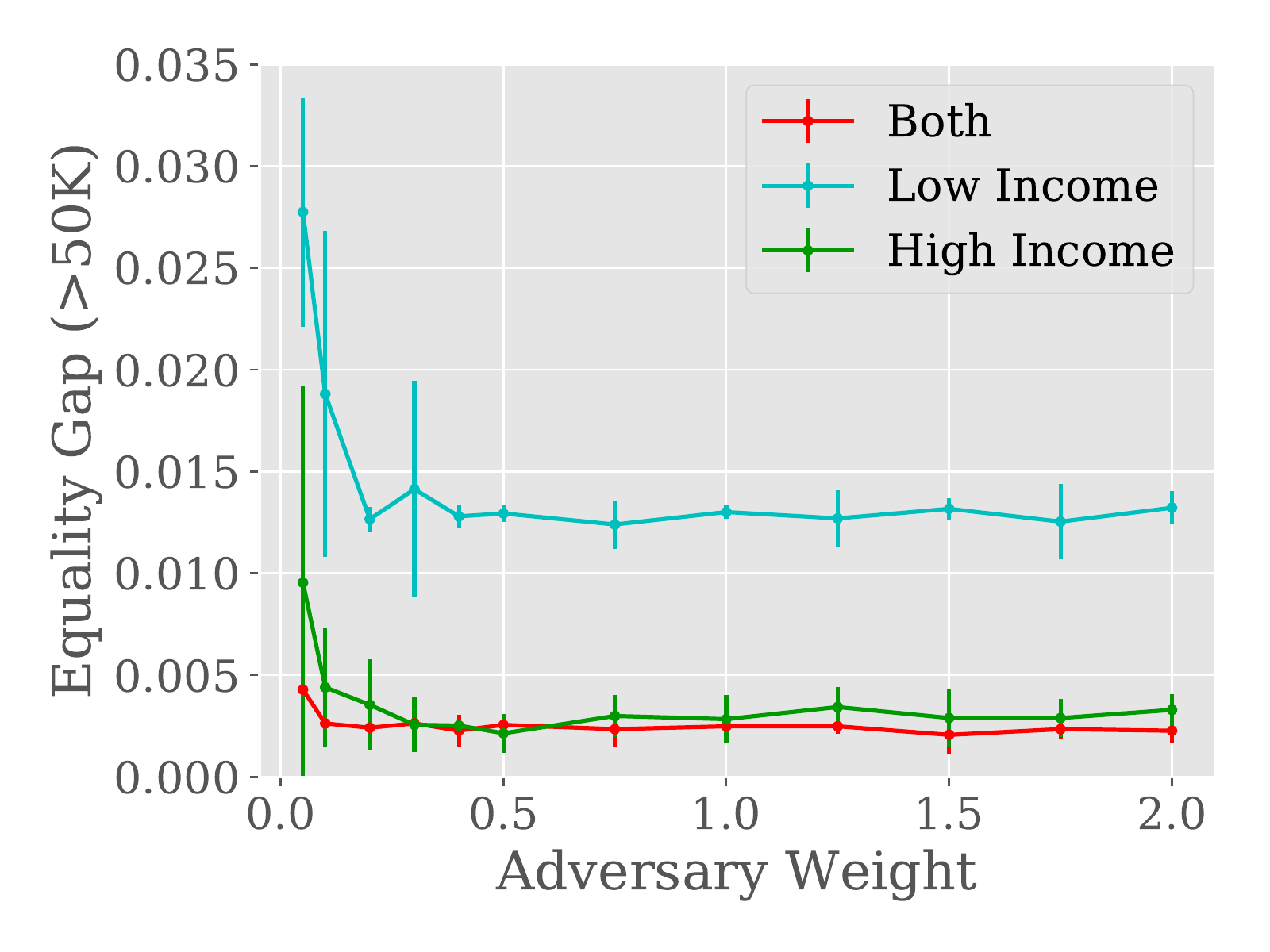}
    \includegraphics[width=0.49\columnwidth]{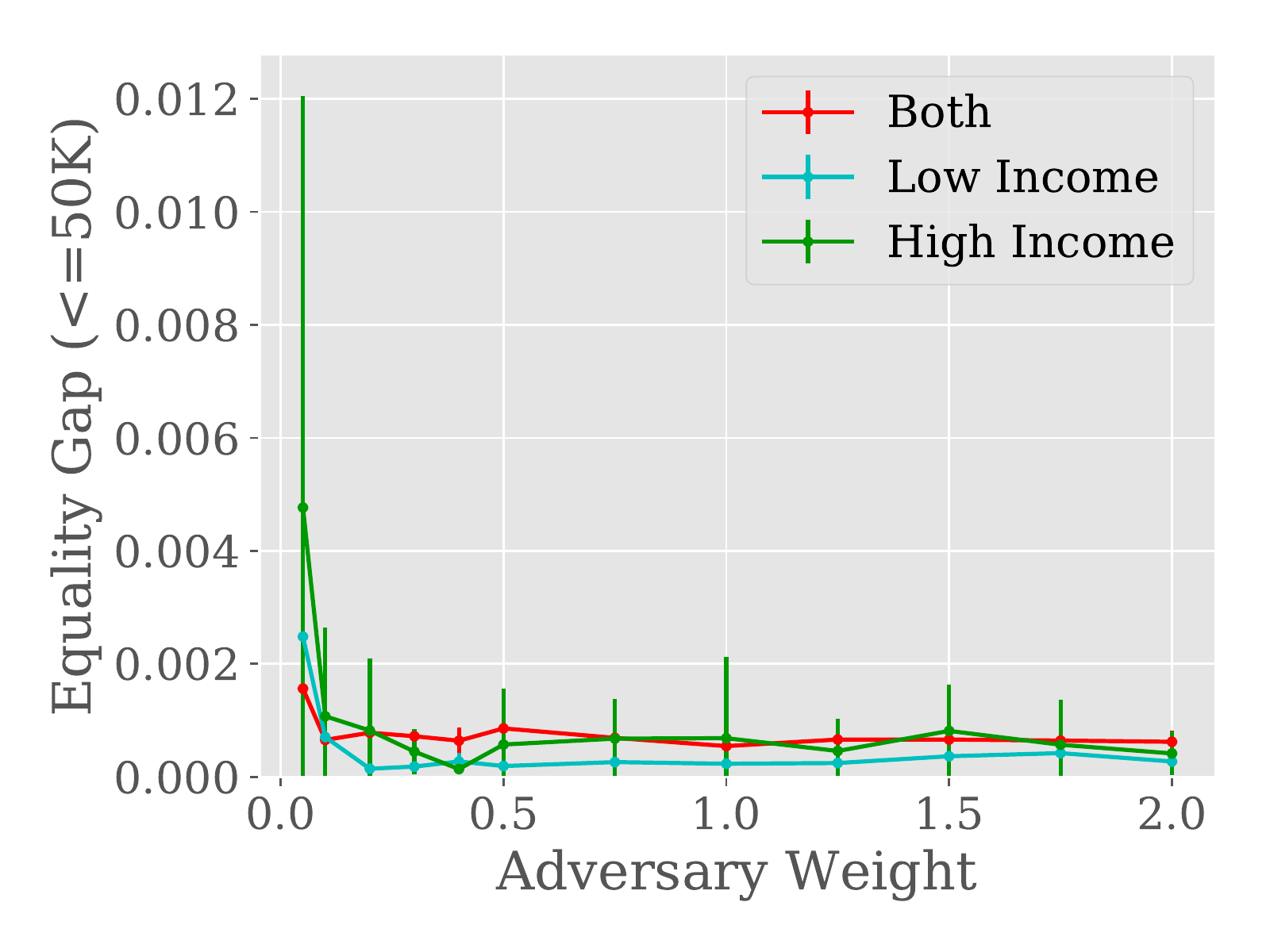}
    \includegraphics[width=0.49\columnwidth]{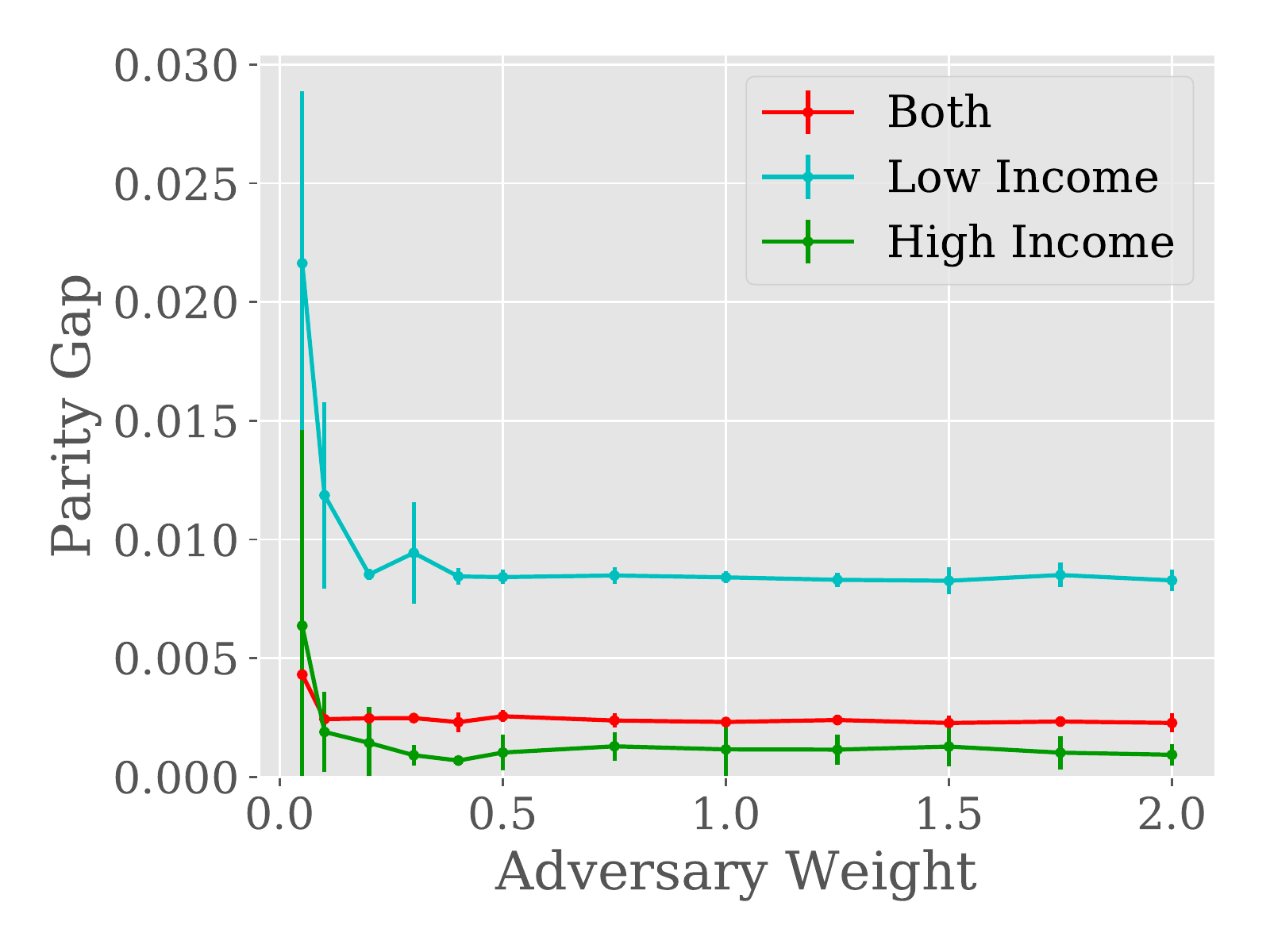}
    \vspace{-2mm}
    \caption{Fairness from different distributions over the primary label $Y$ (while balanced in the sensitive attribute $Z$).}
    \label{fig:primaryskew}
\end{figure*}

\begin{figure*}[tbh]
    \centering
    \includegraphics[width=0.49\columnwidth]{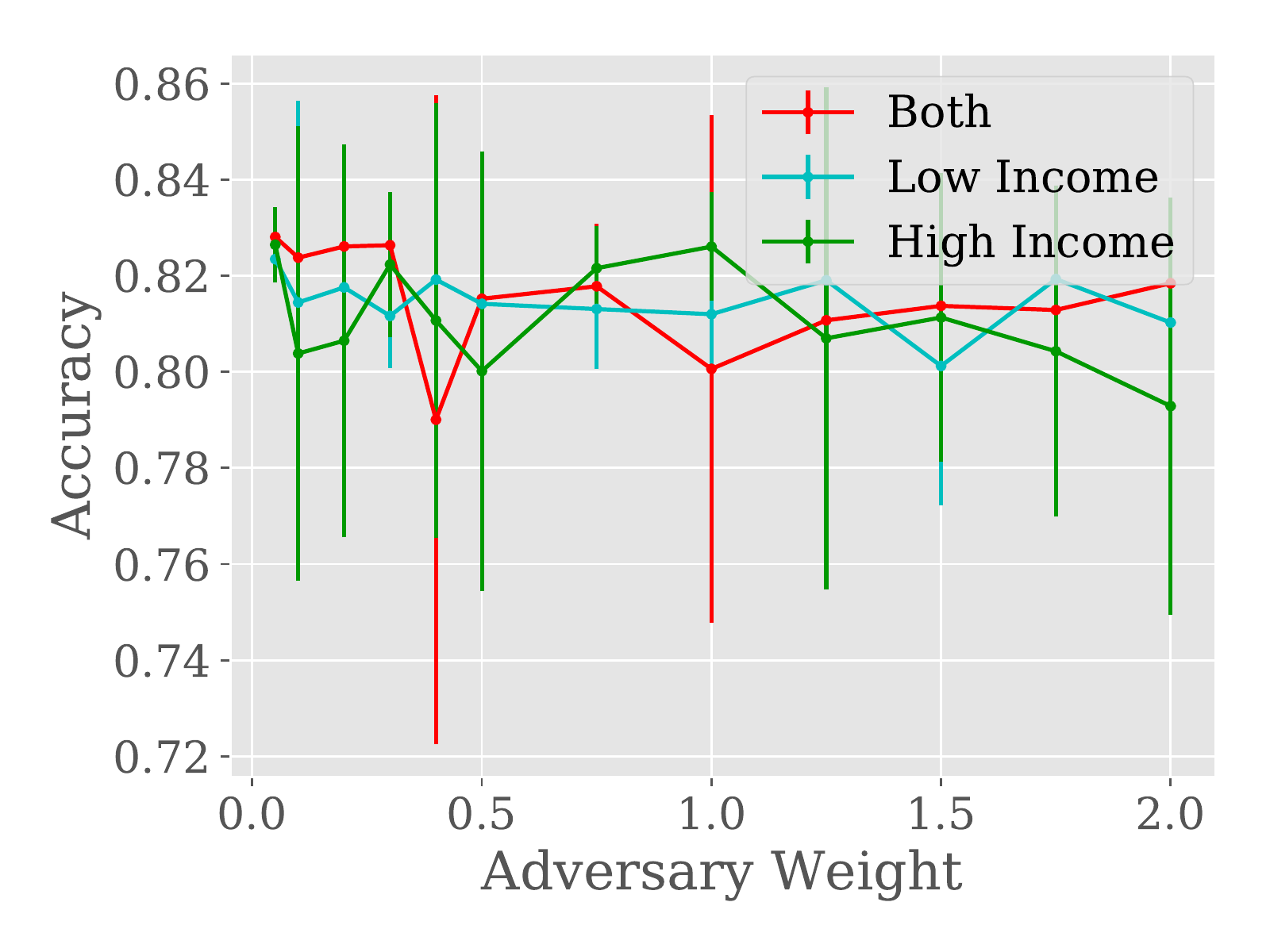}
    \includegraphics[width=0.49\columnwidth]{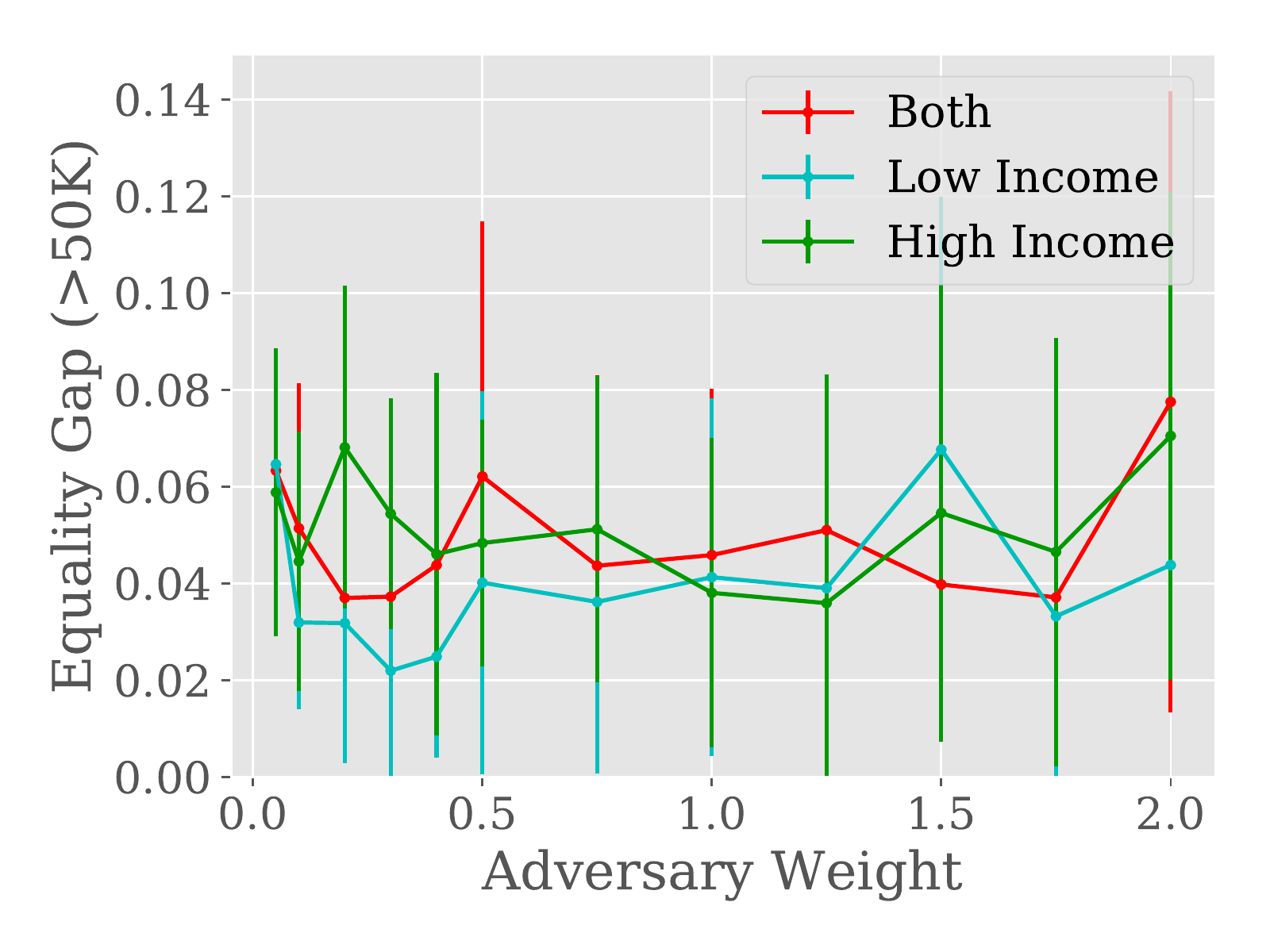}
    \includegraphics[width=0.49\columnwidth]{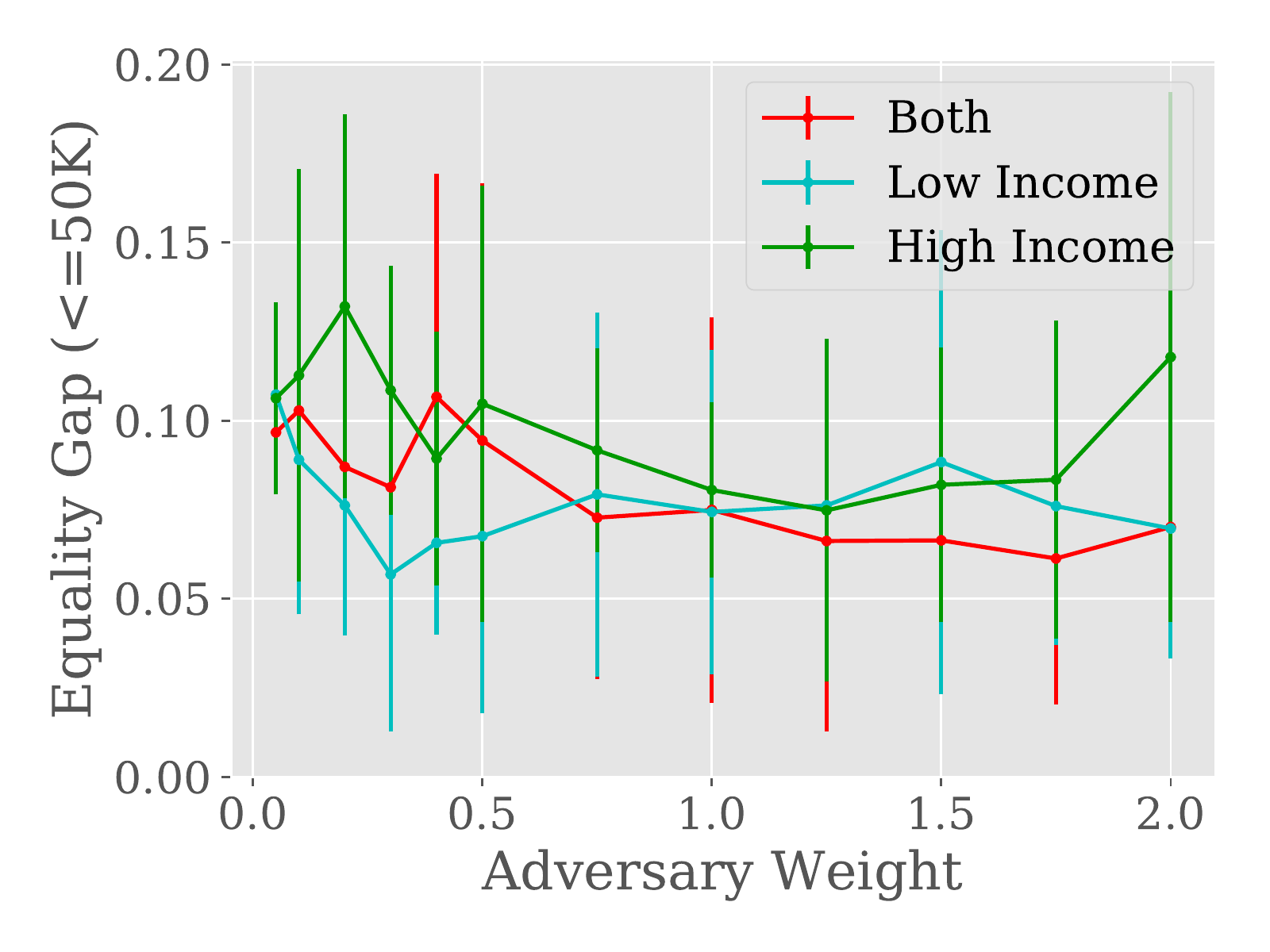}
    \includegraphics[width=0.49\columnwidth]{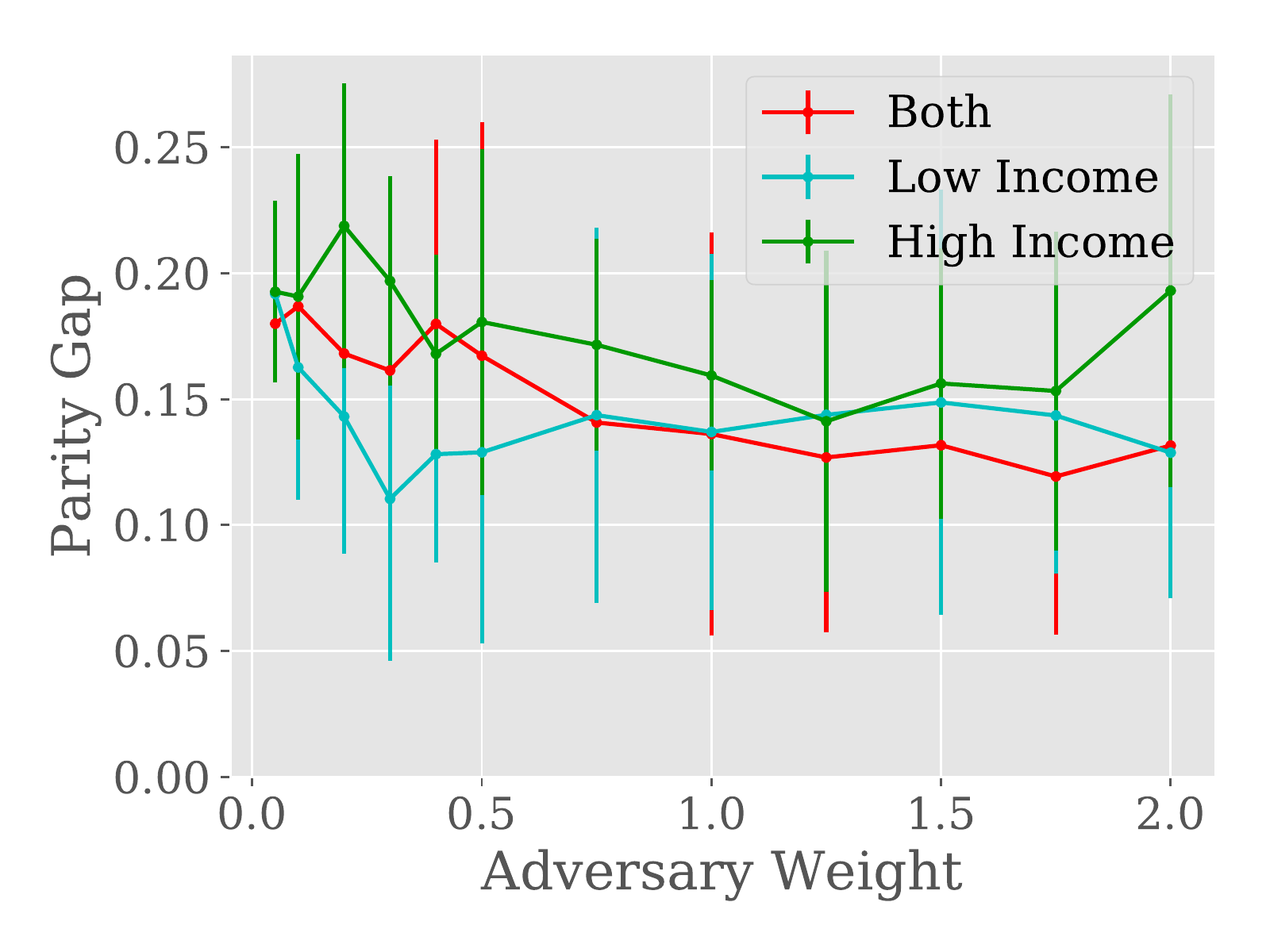}
    \vspace{-2mm}
    \caption{Fairness from different distributions over the primary label $Y$ (while unbalanced in the sensitive attribute $Z$).}
    \label{fig:primaryskew_unbalanced}
\end{figure*}

\begin{figure*}[tbh]
    \centering
    \includegraphics[width=0.49\columnwidth]{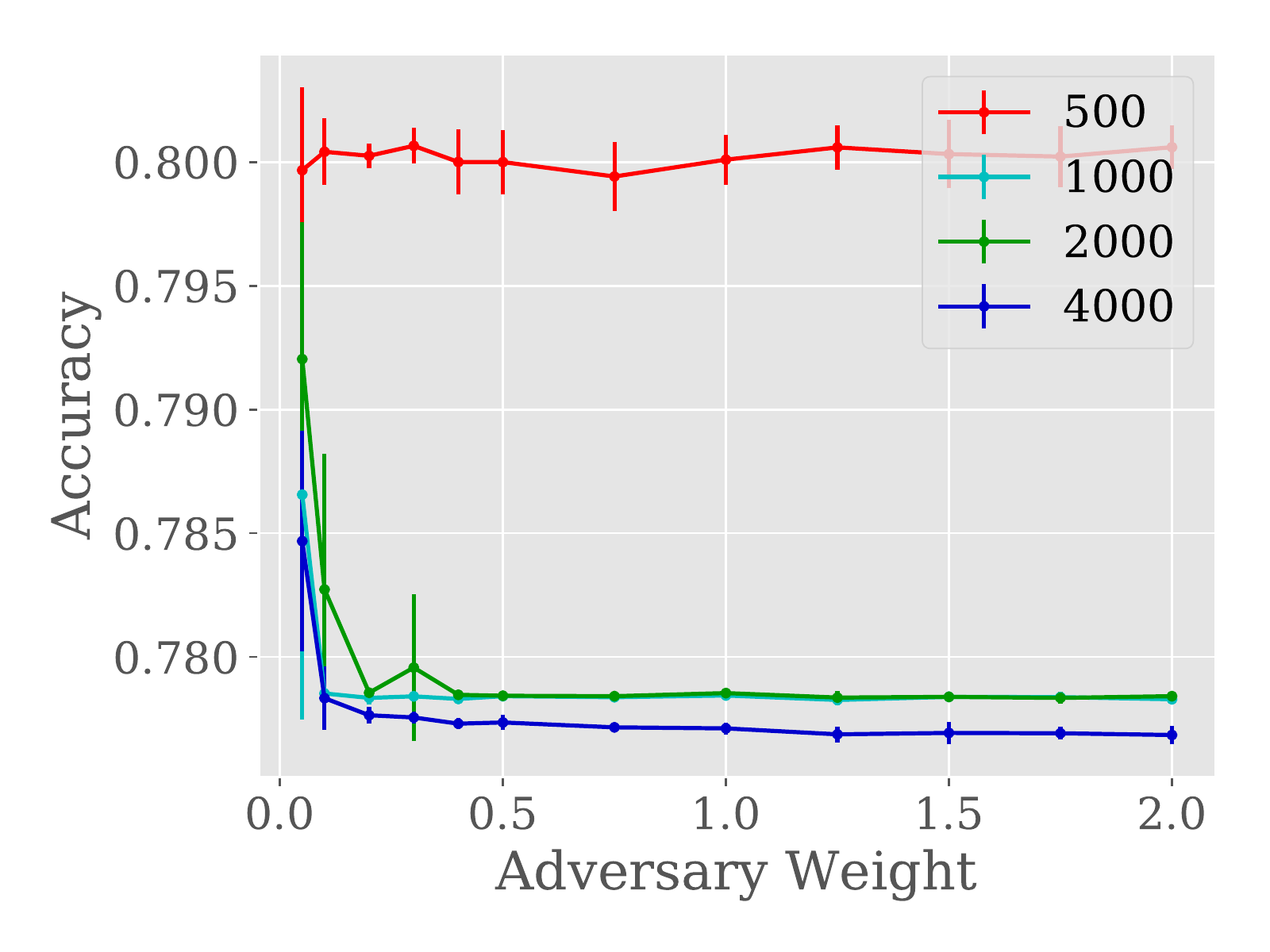}
    \includegraphics[width=0.49\columnwidth]{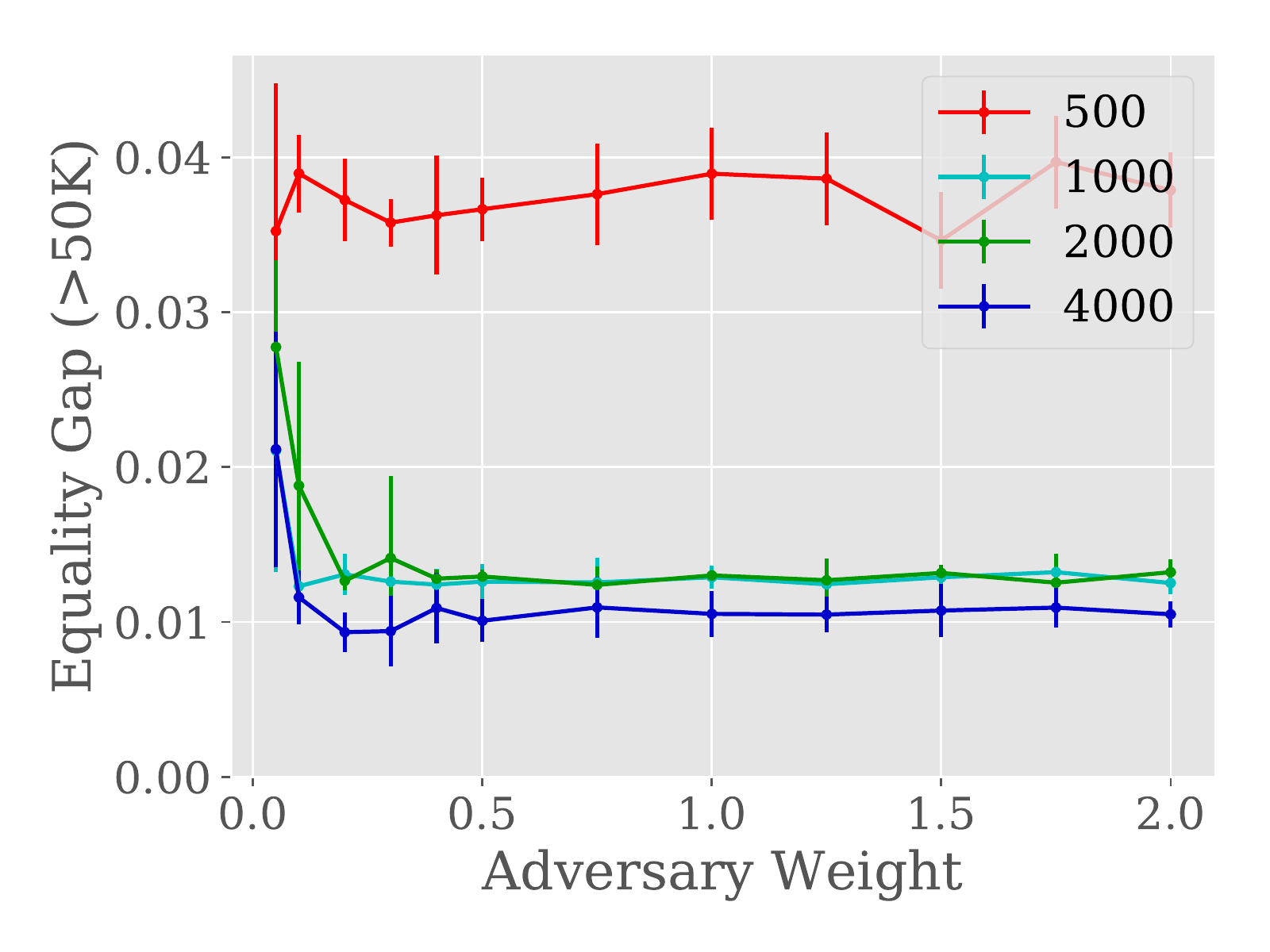}
    \includegraphics[width=0.49\columnwidth]{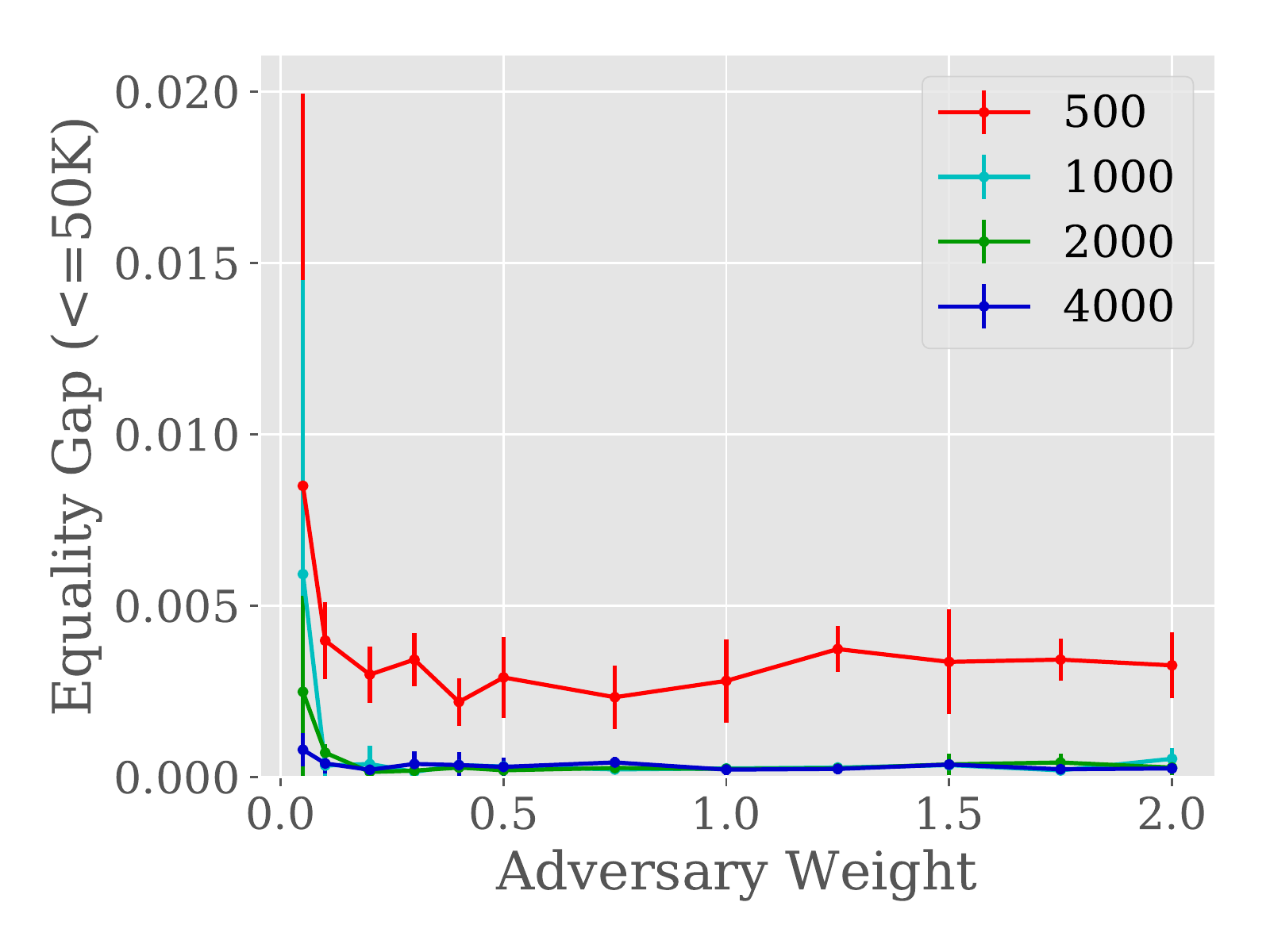}
    \includegraphics[width=0.49\columnwidth]{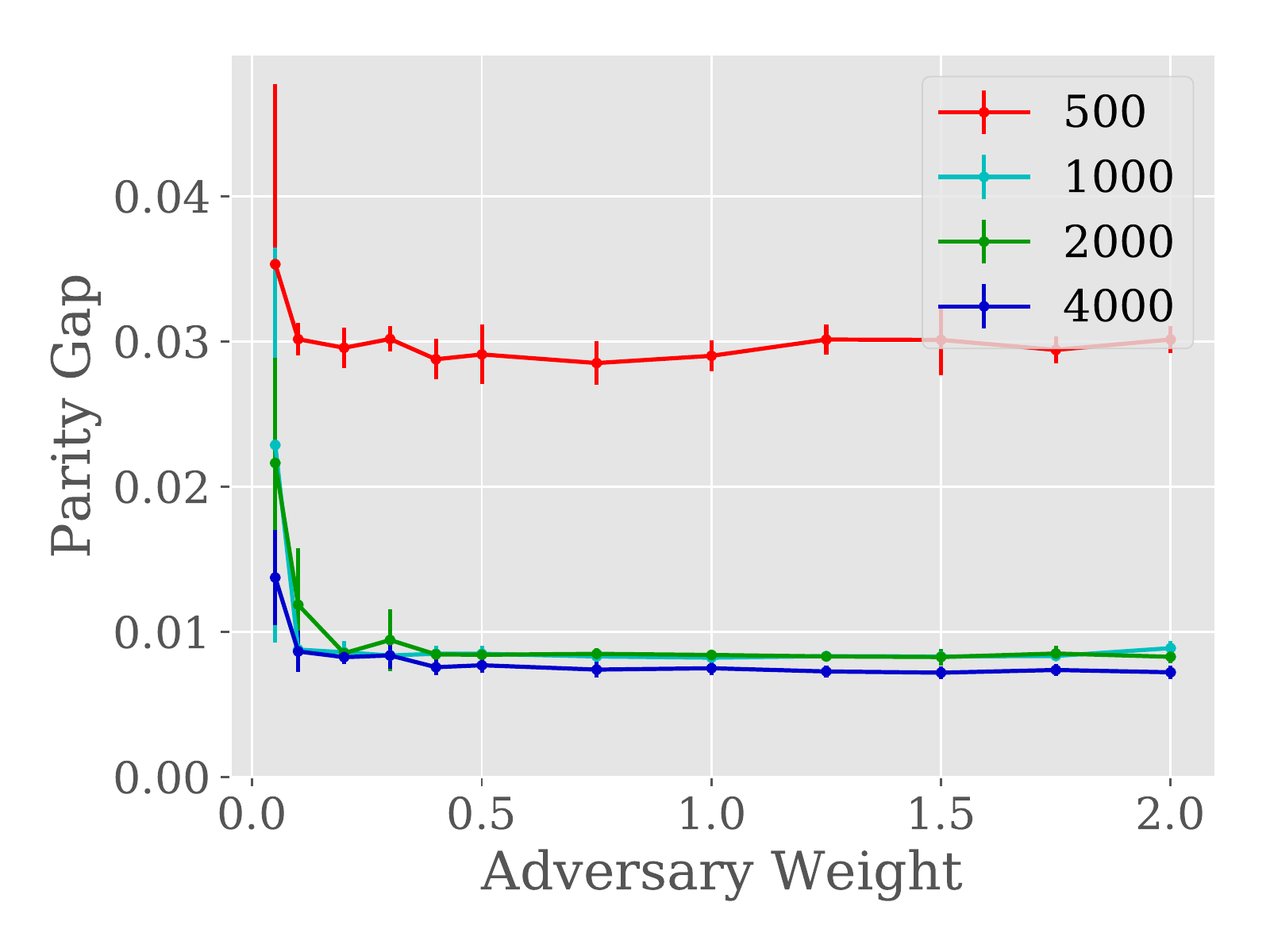}
    \vspace{-2mm}
    \caption{Effect of dataset size (for dataset balanced across gender $Z$ and only including low-income examples).}
    \label{fig:datasize_low}
\end{figure*}

\subsection{Skew in Sensitive Attribute}
One of the most significant findings is that the distribution of examples over the sensitive attribute is crucially important to the performance of the adversary.  We run experiments with both balanced and unbalanced distribution over $Z$.  We show the results for balanced data in Figure \ref{fig:primaryskew} and unbalanced data in Figure \ref{fig:primaryskew_unbalanced}.  As is clear, using balanced data results in a \emph{much} stronger effect from the adversarial training.  Most obviously, we see that the balanced data stabilizes the model, with much smaller standard deviation over results with the exact same training procedure.  Second, we observe that the balanced data much more significantly improves the fairness of the models (across all metrics) but also decreases the accuracy of the model in the process.  

\subsection{Skew in Primary Label}
Next we study the effect of the distribution over the primary label $Y$.  That is, we consider cases where our adversarial head is trained on data exclusively from users with low income ($\leq$50K), high income ($>50$K) or an equal balance of both.  As was described in Section \ref{sec:datadef}, these different distributions theoretically align with different definitions of ``fairness.''  As can be seen in Figure \ref{fig:primaryskew}, we find that different distributions give significantly different results.  Matching the theory in Section \ref{sec:datadef}, we find that the using data from high income users most helps improve equality of opportunity for the high income label, and using data from low income users most helps improve equality of opportunity for the low income label.  Using data from both groups helps on all metrics.

\subsection{Amount of Data}
Additionally, we explore how much data on the sensitive attribute is necessary to improve the fairness of the model.  We vary the size of the dataset and observe the scale of the effect on the desired metrics.  
In most cases, even using only 500 examples (1.5\% of the training data) has a significant effect on the fairness metrics. 
We show in Figure \ref{fig:datasize_low} one of the more conservative cases.
Here, when testing with only low-income, gender-balanced samples, we still observe a strong effect with relatively small datasets.  This is especially encouraging for cases where the sensitive attribute is expensive observe as even a small sample of that data is useful.

\section{Discussion}
This work is motivated by the common challenges in observing sensitive attributes during model training and serving.  We find a mixture of encouraging and challenging results.  Encouragingly, we find that even small samples of adversarial examples can be beneficial in improving model fairness.  Additionally, although it may require more time or more complex techniques, we find that having a balanced distribution of examples over the sensitive attribute significantly improves fairness in the model.  

The empirical results here are also interesting relative to previous theoretical results.  Where as \cite{hardt2016equality} focuses on equality of outcomes, this method encourages unbiased latent representations \emph{inside} the model.  This appears to be a stronger condition if enforced exactly, which can be good for ensuring fairness but possibly harming model accuracy. 
In practice, we have observed that a more sensitive tuning of $\lambda$ finds more amenable trade-offs.

\section{Conclusion}
In this work we have explored the effect of data distributions during adversarial training of ``fair'' models.
In particular, we have mode the following contributions:
\begin{enumerate}
    \item We connect the varying theoretical definitions of fairness to training procedures over different data distributions for adversarially-trained fair representations.
    \item We find that using a balanced distribution over the sensitive attribute for adversarial training is much more effective than a random sample. 
    \item We empirically demonstrate the connection between the adversarial training data and the fairness metrics.
    \item We observe that remarkably small datasets for adversarial training are effective in encouraging more fair representations.
\end{enumerate}

\vspace{1mm}
{\small
\noindent \textbf{Acknowledgements.} We would like to thank Charlie Curry, Pierre Kreitmann, and Chris Berg for their feedback leading up to this paper.
}

\newpage
\bibliographystyle{ACM-Reference-Format}
\bibliography{alex} 

\end{document}